\documentclass[lettersize,journal]{IEEEtran}
\usepackage{amsmath,amsfonts}
\usepackage{algorithmic}
\usepackage{algorithm}
\usepackage{array}
\usepackage[caption=false,font=normalsize,labelfont=sf,textfont=sf]{subfig}
\usepackage{textcomp}
\usepackage{stfloats}
\usepackage{url}
\usepackage{verbatim}
\usepackage{graphicx}
\usepackage{cite}
\usepackage{booktabs} 
\begin{document}

\title{Solving the Clustering Reasoning Problems by Modeling a Deep-Learning-Based Probabilistic Model}

\author{Ruizhuo Song, Member, IEEE,  Beiming Yuan
\thanks{This work was supported by the National Natural Science Foundation of China under Grants 62273036. Corresponding author: Ruizhuo Song, ruizhuosong@ustb.edu.cn}
\thanks{Ruizhuo Song and Beiming Yuan are with the Beijing Engineering Research Center of Industrial Spectrum Imaging, School of Automation and Electrical Engineering, University of Science and Technology Beijing, Beijing 100083, China (Ruizhuo Song email: ruizhuosong@ustb.edu.cn and Beiming Yuan email: d202310354@xs.ustb.edu.cn). }

\thanks{Ruizhuo Song and Beiming Yuan contributed equally to this work.}
}

\markboth{Journal of \LaTeX\ Class Files,~Vol.~14, No.~8, August~2021}%
{Shell \MakeLowercase{\textit{et al.}}: A Sample Article Using IEEEtran.cls for IEEE Journals}

\IEEEpubid{0000--0000/00\$00.00~\copyright~2021 IEEE}

\maketitle

\begin{abstract}

Visual abstract reasoning problems pose significant challenges to the perception and cognition abilities of artificial intelligence algorithms, demanding deeper pattern recognition and inductive reasoning beyond mere identification of explicit image features. Research advancements in this field often provide insights and technical support for other similar domains. In this study, we introduce PMoC, a deep-learning-based probabilistic model, achieving high reasoning accuracy in the Bongard-Logo, which stands as one of the most challenging clustering reasoning tasks. PMoC is a novel approach for constructing probabilistic models based on deep learning, which is distinctly different from previous techniques.
PMoC revitalizes the probabilistic approach, which has been relatively weak in visual abstract reasoning. 

\end{abstract}


\section{Introduction}
\IEEEPARstart{D}{eep} neural networks have achieved remarkable success in various fields, including computer vision \cite{ImageNet,AlexNet,ResNet}, natural language processing \cite{Transformer, Bert, GPT-3}, generative models \cite{GAN,VAE,DiffusionModel}, visual question answering \cite{VQA,CLEVERdataset}, and abstract reasoning\cite{RPM,Bongard1,Bongard2}. Deep learning, as a pivotal branch of machine learning, simulate the learning process of the human brain by establishing multilayered neural networks, thereby facilitating the learning and inference of intricate patterns within data\cite{Human-in-the-loop}. Deep learning has found extensive applications to address diverse and intricate pattern recognition and reasoning problems\cite{RPM,Bongard1,Bongard2}.

Visual abstract reasoning problems occupy a pivotal position in the field of artificial intelligent. They involve identifying and comprehending hidden patterns, structures, and relationships within graphics, reflecting a deep cognitive process. Research on these problems has propelled advancements in artificial intelligent technology, fostering continuous optimization of key techniques such as feature extraction and modeling. The outcomes of this research have broad applications, spanning security, healthcare, and autonomous driving, among others. Furthermore, the capability of effective reasoning is a significant indicator of the intelligence level of AI systems. Addressing these problems poses challenges but also presents tremendous opportunities for breakthroughs, especially with the advent of deep learning techniques\cite{learning system}. In summary, visual abstract reasoning problems are not only crucial to basic research but also hold the potential to drive technological progress and expand the application scope of artificial intelligent.

Visual abstract reasoning problems typically come in two forms: clustering reasoning and progressive pattern reasoning. 
For instance, Ravens Progressive Matrices (RPM)\cite{RPM} serves as a benchmark for progressive pattern reasoning problems, while Bongard problems\cite{Bongard1,Bongard2} stand as a paragon for clustering inference, both encompassing learning requirements that span from perception to reasoning.

\subsection{Bongard-logo database}
Bongard problems \cite{Bongard1}, contrasting with RPM (Raven's Progressive Matrices) problems, are noteworthy exemplars of small-sample learning challenges. These problems typically present a series of images divided into two distinct groups: primary and auxiliary. The primary group comprises images that adhere to a specific set of rules defining an abstract concept, whereas the auxiliary group includes images that deviate from these rules in varying degrees. The intricacy of Bongard problems lies in the requirement for deep learning algorithms to precisely categorize ungrouped images into their corresponding groups based on subtle pattern recognition and abstract reasoning.

\begin{figure}[htp]\centering
	\includegraphics[trim=8cm 0cm 6cm 0cm, clip, width=8.5cm]{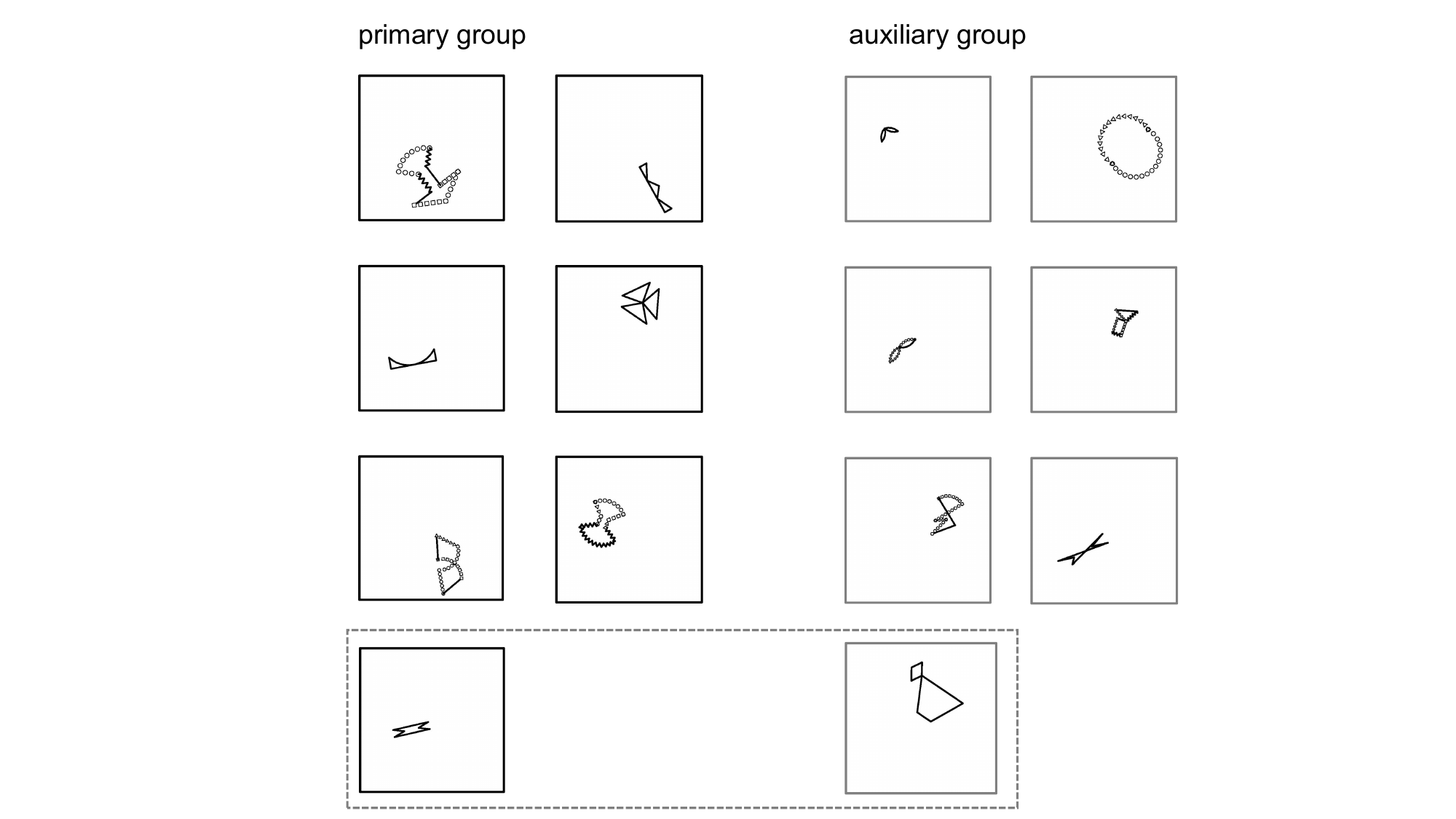}
	\caption{Bongard case}
\label{Bongard case}
\end{figure}

Within the domain of abstract reasoning, Bongard-logo\cite{Bongard2} problems serve as a particular instantiation of Bongard problems, posing a significantly high level of reasoning difficulty. Each Bongard-logo problem consists of 14 images, with 6 images belonging to the primary group, 6 to the auxiliary group, and the remaining 2 serving as categorization options. These images are comprised of various geometric shapes, and the arrangements of these shapes serve as the basis for grouping. Figure \ref{Bongard case} illustrates an exemplary Bongard-logo problem, showcasing the complexity and abstract nature of these challenges.

In figure \ref{Bongard case}, each Bongard problem is comprised of two distinct image sets: primary group A and auxiliary group B. Group A contains 6 images, with the geometric entities within each image adhering to a specific set of rules. Conversely, group B includes 6 images that deviate from the rules established in group A. The task at hand requires determining whether the images in the test set align with the rules dictated by group A. The level of difficulty associated with these problems varies depending on their structural complexity.

Furthermore, Bongard-logo problems are classified into three distinct conceptual categories: 1) Free Form problems (FF), in which each shape is composed of randomly sampled action strokes, potentially resulting in one or two shapes per image; 2) Basic Shape problems (BA), corresponding to the identification of a single shape category or combinations of two shape categories represented within given shape patterns; and 3) Abstract Concepts (HD), designed to assess a model's proficiency in discovering and reasoning about abstract concepts such as convexity, concavity, and symmetry. These categorizations underscore the breadth and depth of reasoning abilities required to tackle Bongard-logo problems effectively.

\section{Related work}

This section is devoted to comprehensively presenting the significant research achievements related to the RPM (Raven's Progressive Matrices) problem and the Bongard-Logo problem, as derived from past studies. These notable research milestones have played a pivotal role in driving forward the advancements in the field of artificial intelligence, specifically in relation to abstract reasoning capabilities. By thoroughly examining these milestones, we aim to foster a deeper understanding of the complexities and nuances associated with abstract reasoning challenges. Moreover, we hope to provide invaluable insights and guidance that can serve as a springboard for future research efforts, ultimately contributing to the continued growth and development of AI capabilities in tackling abstract reasoning tasks.







\subsection{Bongard solver}

In recent years, three main approaches have been employed to address Bongard problems: methods based on language feature models, methods based on convolutional neural network models, and generated datasets.

Methods based on language feature models\cite{Bongard1}: somes researchers proposed a solution leveraging image-based formal languages, transforming image information into symbolic visual vocabulary, and subsequently tackling BP problems via symbolic languages and Bayesian inference. However, the limitation of this approach lies in its inability to directly apply to complex, abstract conceptual BP problems due to the need for reconstructing symbolic systems.

Methods based on convolutional neural network models\cite{Bongard3}: some researchers constructed an image dataset consisting of simple shapes and utilized these images for pre-training to cultivate a feature extractor. Subsequently, image features from Bongard problems were extracted for image classification, determining whether test images conformed to the rules. Yun adopted a similar approach, initially pre-training with images containing visual features from BP problems to extract BP image features and then integrating an additional classifier for discrimination.

Methods based on Generated datasets\cite{Bongard2}: some researchers employed fundamental CNNs, relation networks like WReN-Bongard, and meta-learning techniques, but their performance on the Bongard
-Logo database was less than optimal.

These represent the main solutions targeted at Bongard problems in recent years, each with its own strengths and limitations.



\subsection{Transformer and Vision Transformer (ViT)}
The Transformer \cite{Transformer} model, originally designed for machine translation tasks, is a encoder-decoder architecture that relies heavily on attention mechanisms, particularly self-attention. This mechanism enables the model to simultaneously consider other words in a sentence while processing a specific word, effectively capturing contextual relationships between them. On the other hand, Vision Transformer (ViT) \cite{ViT} represents the application of the Transformer model in the image domain. It processes image data by dividing it into small patches, linearly embedding each patch into a vector, and then leveraging attention mechanisms to capture global contextual information for image recognition and processing. Both Transformer and Vision Transformer utilize self-attention mechanisms and encoder-decoder architectures, but they differ in the types of data they handle and the manner in which they process it.

\subsection{Sinkhron distance}

The Sinkhorn distance\cite{Sinkhorn} is a notion derived from the field of optimal transport theory, offering a computationally efficient alternative to the classical Wasserstein distance\cite{Wasssertein}. At its core, it approximates the cost of transporting mass from one probability distribution to another by relaxing the constraint of mass preservation through a regularization term. This regularization, typically parameterized by a scalar value known as the entropy regularization parameter, allows for a smoother, more tractable optimization problem.


\subsection{Discrete-Time Inference and Establishment of Contextual Probability Models}
The theory of isochronous constraint reasoning \cite{Isoperimetric constraint inference} has brought profound insights to the design of reasoning problem solvers. By integrating logical reasoning from this theory, along with optimization strategies and discrete-time system analysis, we can more precisely understand and predict complex logical relationships in reasoning problem. Additionally, drawing on the experience of isochronous constraint reasoning in addressing nonlinear problems, and adopting data-driven techniques, we can build an efficient and accurate reasoning problem solver\cite{Data-Driven}, effectively tackling challenges in visual entities and elements sequence reasoning.

The contextual analysis\cite{context} method provides significant insights into the design of abstract reasoning problem solvers. By deeply analyzing the contextual relationships between graphics, the solver can more accurately grasp the logical connections and patterns of change among them. Specifically, this method urges consideration of the position and role of graphics in the overall sequence, leading to a more comprehensive understanding of each graphic's significance. Encouraging a focus on the mutual influence among graphics rather than viewing each in isolation, the solver's design benefits from capturing and utilizing contextual information, thereby enhancing problem-solving accuracy and efficiency, especially when dealing with complex and nonlinear graphic relationships. In conclusion, the contextual analysis approach serves as a powerful thinking tool for the design of visual abstract reasoning problem solvers, facilitating the development of smarter and more efficient algorithms.

Attention-based visual models\cite{mini pose} demonstrate significant superiority by focusing on critical information within images, thereby enhancing processing efficiency and accuracy. These models enable a sharper focus on pertinent data while adapting to multi-modal data relationships, resulting in their widespread application across various domains. It also serves as the underpinning of our methodological approach.

In light of the challenges and opportunities associated with modeling system problem spaces using probabilistic models\cite{probabilistic}, this paper opts to leverage deep learning algorithms to construct a deep-learning-based probabilistic model aimed at addressing visual abstract reasoning problems.

\section{Methodology for Bongard-Logo}

The Bongard-logo\cite{Bongard1} and Raven's Progressive Matrices (RPM)\cite{RPM} tasks are both paradigms of abstract reasoning that require the extractor's ability in unveiling precise concepts hidden within abstract representations. These concepts are indicative of more elevated abstractions, surpassing rudimentary ideas such as ``pixel configuration patterns". Frequently, these elevated abstractions encapsulate diverse degrees of human-centered pre-existing knowledge knowledge, including elements like form, dimension, hue, spatial orientations, concave and convex contours of shapes, and shape completion. Using distinct querying approaches, the Bongard-logo and RPM tasks endeavor to evaluate the extractor's expertise in unveiling and assimilating these abstract concepts.

The methodology presented in this section targets clustering reasoning problems within visual abstract reasoning, attempting to solve them by modeling a probabilistic model. The Bongard-Logo problem stands as a renowned example of clustering reasoning, drawing significant attention due to its exceptional reasoning difficulty and the current lack of progress in solving it. This paper aims to achieve promising results on the Bongard-Logo problem, demonstrating the potential of probabilistic models in solving clustering reasoning tasks.

\subsection{A method for solving the bongard-logo problem based on sinkhorn distance (SBSD)}
In the Bongard-Logo problem, we denote $x_{ij}$ as the $j$-th image in the $i$-th Bongard-Logo problem, where $i\in [1,n]$ signifies the question number, with $n$ being the total number of questions. Specifically, $\{x_{ij}\}_{j=1}^6$ represents images in the $i$-th primary group, while $\{x_{ij}\}_{j=8}^{13}$ represents images in the $i$-th auxiliary group. Additionally, $x_{i7}$ represents the test image to be potentially assigned to the $i$-th primary group, and $x_{i14}$ represents the test image to be potentially assigned to the $i$-th auxiliary group.
we let the distribution of the primary group (positive examples) within a Bongard-Logo question be denoted as $p_i(x|y)$, and the distribution of the auxiliary group (negative examples) be denoted as $q_i(x|y)$. Here, y represents the reasoning type condition of the question, where $y \in \{ff, ba, hd\}$.
We are committed to developing a deep learning algorithm that can induce distributions $p_i(x|y)$ and $q_i(x|y)$ with low cross-measure between them.

Directly deriving the image distributions $p_i(x|y)$ and $q_i(x|y)$ of Bongard-Logo without dimensionality reduction and computing the cross measure between these two distributions is quite challenging. Moreover, such an approach would also result in a waste of resources and time. Therefore, we aim to devise a deep learning algorithm, denoted as $f_\theta (z|x)$, which can map samples $x_{ij}$ to latent variables $z_{ij}$. We hope that $f_\theta (z|x)$ can serve as a distribution transformation function, and by measuring the distribution of the latent representation, we can address the Bongard-Logo problem. 
Accordingly, the key objective is to ensure that the encoder $f_\theta (z|x)$ possesses a specific image encoding capability, which necessitates its ability to produce encoded latent variables that demonstrate specific characteristics. Specifically, the distribution of latent variables within the primary group (represented as $p _i'(z|y)$) and the distribution of latent variables within the auxiliary group (represented as $q_i'(z|y)$) should exhibit very low cross-measure.
The Kullback-Leibler (KL) divergence, a measure used to quantify the similarity between distributions, can be employed as an optimization objective for $f_\theta (z|x)$\cite{VAE}. The formula for calculating KL divergence is provided below.
\begin{equation}
    KL(P||Q)=\sum P(x) \log \left(\frac{P(x)}{Q(x)}\right)
\end{equation}
However, due to the characteristics of small sample learning problems, it is challenging to estimate the distributions $p_i'(z|y)$ and $q_i'(z|y)$ of Bongard-Logo image representations. Estimating the specific forms of distributions $p_i'(z|y)$ and $q_i'(z|y)$ poses an even greater challenge. Additionally, the limitations of the Kullback-Leibler (KL) divergence make it difficult to directly optimize for these conditions and obtain a highly effective deep model.

Hence, we endeavored to utilize baseline models, specifically ResNet18, for the purpose of encoding images within the Bongard-Logo paradigm, denoted as $\{x_{ij}\}_{j=1}^{14}$, into meaningful representations, designated as $\{z_{ij}\}_{j=1}^{14}$. These representations, in the context of a Bongard-Logo question, can be interpreted as samples drawn from the distributions $p_i'(z|y)$ and $q_i'(z|y)$. Subsequently, constraints were imposed on the Sinkhorn distance between $p_i'(z|y)$ and $q_i'(z|y)$ through these representations. This methodology was designed primarily because the Sinkhorn algorithm computes the distance without requiring explicit knowledge of the distributions' specific forms, making it highly suitable for scenarios where the forms of $p_i'(z|y)$ and $q_i'(z|y)$ remain unknown.

Specifically, the representations within the primary group, denoted as $\{z_{ij}\}_{j=1}^7$, were randomly partitioned into two distinct subgroups: $\{z_{it}\}$ and $\{{z}_{i\tilde{t}} \}$. By utilizing the Sinkhorn distance between these subgroups, constraints were imposed to ensure that the representations within the primary group $\{z_{ij}\}_{j=1}^7$ adhered to a consistent, albeit unknown, distribution. Analogously, similar constraints were introduced between the distributions represented by $p_i'(z|y)$ and $q_i'(z|y)$, leveraging representations from both the primary and auxiliary groups to minimize the cross-measure.

\begin{figure}[htp]\centering
	\includegraphics[trim=1cm 0cm 9cm 0cm, clip, width=8.5cm]{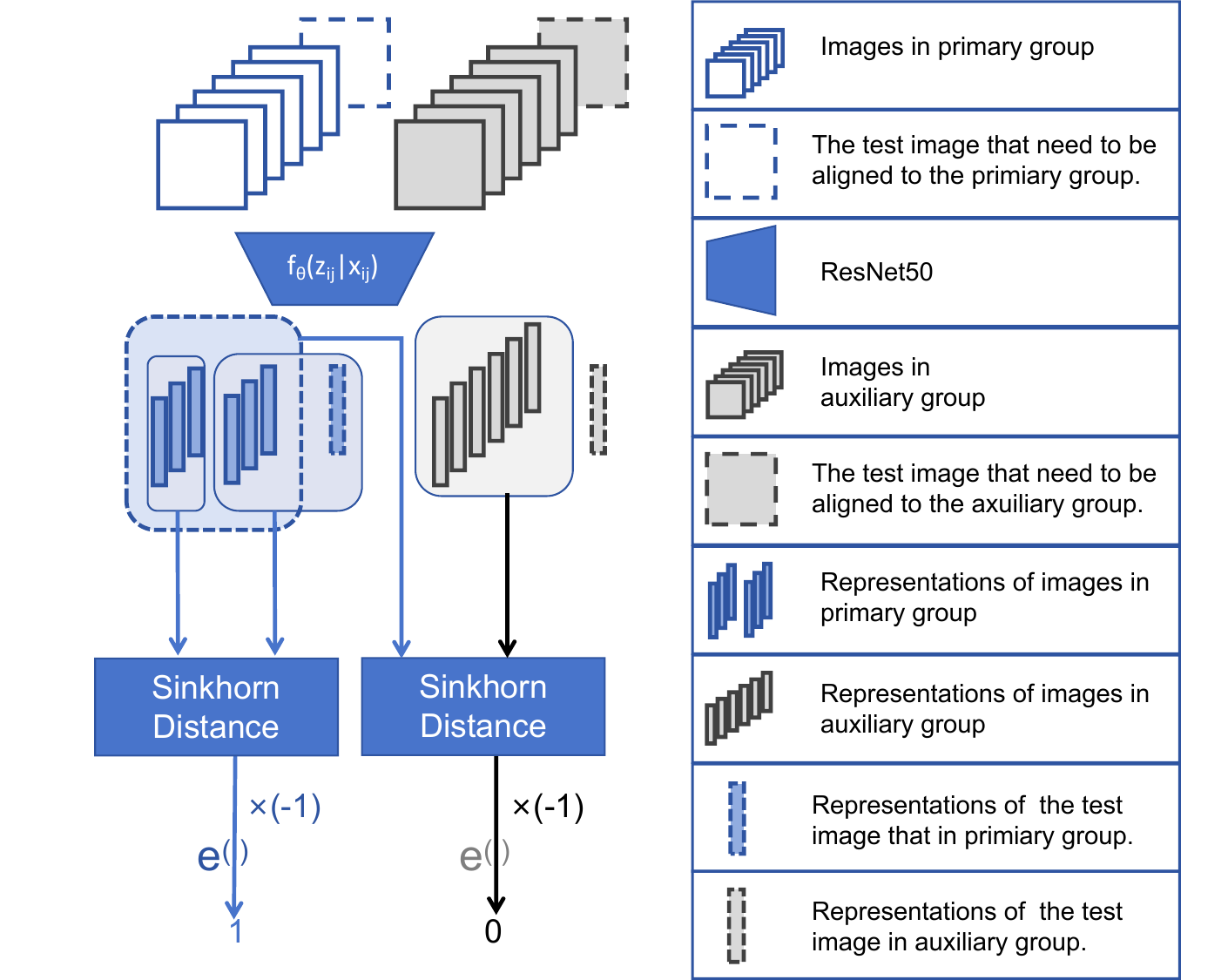}
	\caption{Feedforward process of SBSD}
\label{sinkhorn}
\end{figure}

The aforementioned constraints can be explicitly articulated through a mathematical loss function.
\begin{align}
    &{\ell}(\{z_{ij}\}_{j=1}^{14}) \nonumber\\
    &= - \log \frac{{{e^{-D(\{z_{it}\} ,\, \{{z}_{i\tilde{t}} \})}}}}{{{e^{-D(\{z_{it}\} ,\, \{{z}_{i\tilde{t}} \})}} + {e^{-D(\{z_{ij}\}_{j=1}^{7} ,\, \{{z}_{ij}\}_{j=8}^{14})}} }}
\end{align}
Let $D(a,b)$ denote the calculation of the Sinkhorn distance between the distributions associated with sets $a$ and $b$. Additionally $\{z_{it}\} \cup \{{z}_{i\tilde{t}} \}= \{{z}_{ij} \}_{j=1}^7$ and $\{z_{it}\} \cap \{{z}_{i\tilde{t}} \}= \emptyset
$.
Despite our endeavors, this approach did not yield satisfactory results. We hypothesize that this was primarily due to the limited sample size within a single Bongard-logo problem, which, when collectively treated as sampling outcomes, may not meet the prerequisite sampling requirements of the Sinkhorn distance. An illustrative diagram depicting this process can be found in Figure \ref{sinkhorn}.
 
In summary, SBSD represents our initial and most direct attempt to model the Bongard-Logo problem as a probabilistic framework. SBSD treats the representations of images within the primary group as samples from the primary group distribution $p_i'(z|y)$, and the auxiliary group follows the same approach. By utilizing these samples, the Sinkhorn distance is employed to calculate and represent the distance between the primary $p_i'(z|y)$ and auxiliary distributions $q_i'(z|y)$. This distance can be used to solve the Bongard-Logo problem.

\subsection{PMoC(Probability Model of Concept)}

After conducting experiments, we discovered that SBSD had achieved limited progress in reasoning accuracy. However, this paper aims to make more contributions and breakthroughs. Given the difficulties associated with utilizing the SBSD approach, which involves optimizing deep networks based on the distances between distributions, this paper proposes the design of the PMoC (Probability Model of Concept) network. PMoC explores an alternative approach by shifting the constraint conditions. The training objective shifts from focusing on the deep model's ability to disentangle distribution $p_i'(z|y)$ and distribution $q_i'(z|y)$ to emphasizing its capacity to compute the probability of a sample  belonging to the distribution $p_i'(z|y)$. For this purpose, we assume that $p_i'(z|y)$ follows a multivariate Gaussian distribution. 

From a macroscopic perspective, our PMoC framework consists of two components: module $f(z|x)$ and module $g (\mu,\sigma^2|z)$.
We leverage the $f(z|x)$ module to map Bongard-Logo images into latent representations, and the $g (\mu,\sigma^2|z)$ module to deduce a Gaussian distribution based on the representations of samples from the primary group. Subsequently, we compute the probabilities of representations from the auxiliary group and the test samples belonging to this distribution. Finally, we employ a cross-entropy loss function to constrain these probabilities, optimizing both the mapping capability of the $f(z|x)$ module and the distribution computation ability of the $g (\mu,\sigma^2|z)$ module.

\begin{figure}[htp]\centering
	\includegraphics[trim=0cm 1cm 7.5cm 0cm, clip, width=8.5cm]{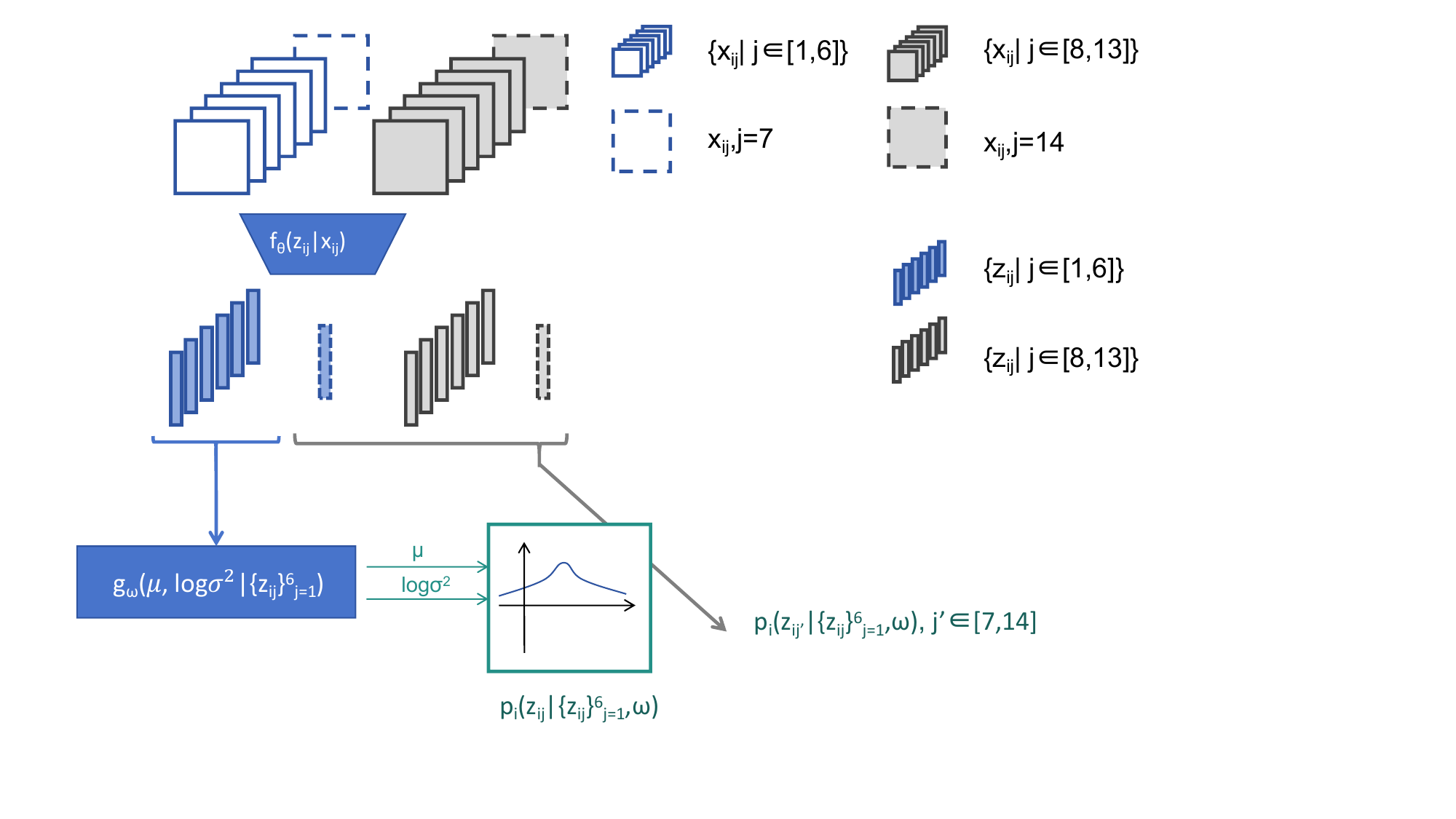}
	\caption{Framework of PMoC.}
\label{Framework}
\end{figure}

Specifically, we utilize a deep network $f_\theta(z_{ij}|x_{ij})$ (serving as the $f(z|x)$ module) to transform the Bongard-Logo images $\{x_{ij}\}^{14}_{j=1}$ into their corresponding latent representations $\{z_{ij}\}^{14}_{j=1}$. Then, another deep network $g_\omega (\mu_i , \sigma_i |\{z_{ij}\}^6_{j=1})$ (acting as the $g (\mu,\sigma^2|z)$ module) is employed to derive an optimizable multivariate Gaussian distribution $p'_{i } (z|\{z_{ij}\}^6_{j=1}, \omega)$ based on the latent representations of the primary group samples $\{z_{ij}\}^6_{j=1}$.
%
%
Next, we compute the probabilities of each latent representation $z_{ij'}$ (where $j'$ ranges from 7 to 14) belonging to this distribution, denoted as the set $\{p'_{i } (z_{ij'}|\{z_{ij}\}^6_{j=1}, \omega)\}_{j'=7}^{14}$.
In essence, the network $f_\theta(z_{ij}|x_{ij})$ functions as an image encoder, mapping input images to their latent representations, while $g_\omega (\mu_i , \sigma_i |\{z_{ij}\}^6_{j=1})$ serves as a logical distribution fitter, calculating the distribution of latent representations and modeling it as a Gaussian distribution. To this end, we adopt a Convolutional Neural Network (CNN) to fit $f_\theta(z_{ij}|x_{ij})$ and a Transformer-Encoder to fit $g_\omega (\mu_i , \sigma_i |\{z_{ij}\}^6_{j=1})$. Further details regarding the design are provided below. This framework is visualized in Figure \ref{Framework}.

\begin{figure}[htp]\centering
	\includegraphics[width=8.5cm]{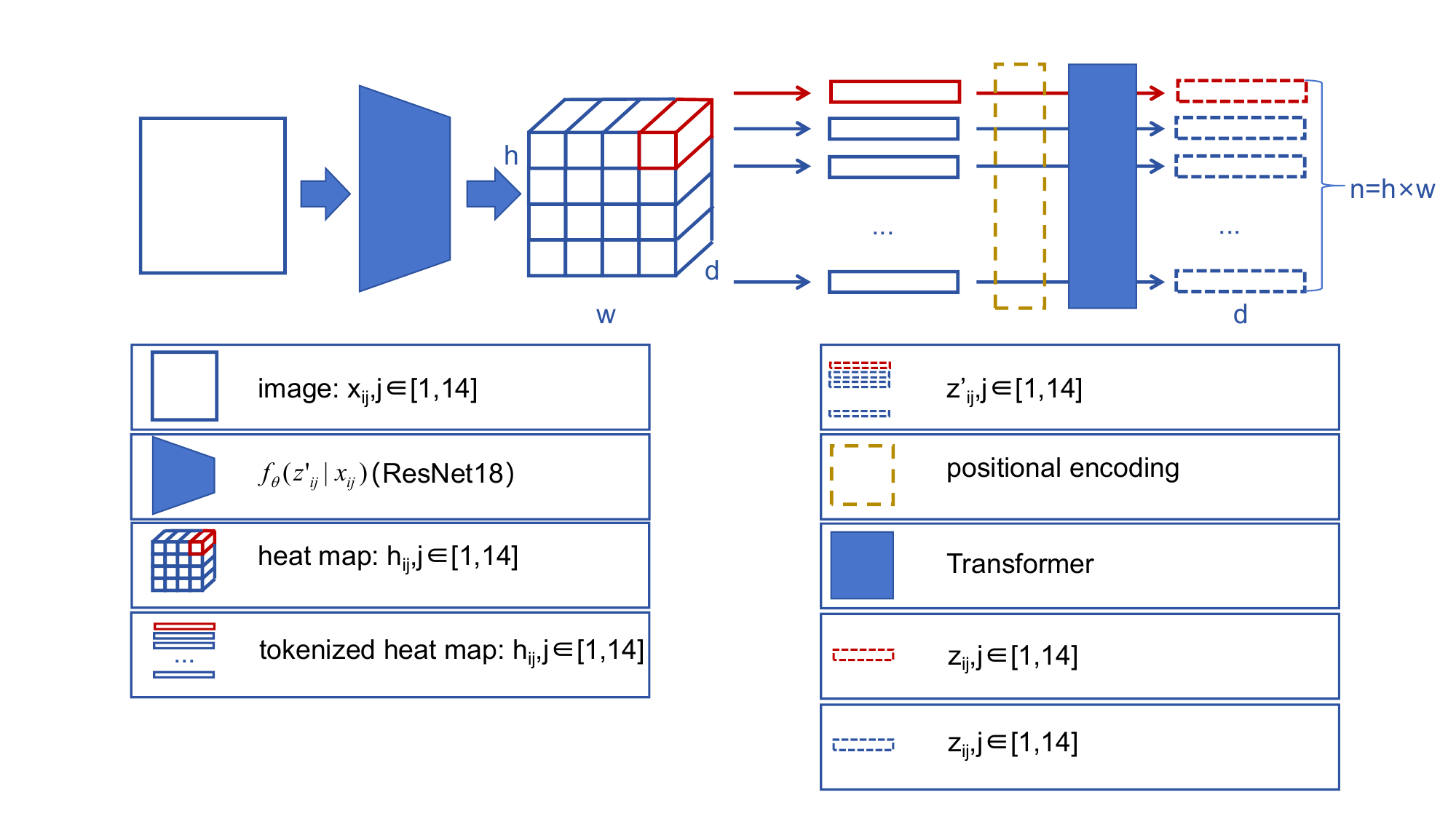}
	\caption{Feedforward process of $f_\theta(z_{ij}|x_{ij})$}
\label{f_z}
\end{figure}

Regarding the network implementation of $f_\theta(z_{ij}|x_{ij})$, we present an approach that combines ResNet18 with Transformer-Encoder, employing this integration as the architecture for the function $f_\theta(z_{ij}|x_{ij})$. In detail, we encode the Bongard-Logo problem images $x_{ij}$ into feature maps $h_{ij}\in R^{h\times w\times d}$ using ResNet18. Subsequently, we calculatethe self-attention results $z'_{ij}\in R^{n\times d}$ among all receptive fields within the feature maps $h_{ij}$. This computation enables the $n$ attention outputs $z'_{ij}$ to capture global information of the image in varying degrees, ensuring that each output retains some level of contextual understanding. We encoded Bongard-Logo images from multiple scales and perspectives using the aforementioned method. We denote  each individual perspective within $z'_{ij}$ as $z_{ij} \in R^{d}$. Since $g_\omega (\cdot |\cdot)$ in this paper treats each perspective within $z'_{ij}$ equally and in parallel, we do not assign different notations to individual perspectives within $z'_{ij}$ and uniformly denote them as $z_{ij}$. The feedforward process of the network $f_\theta(z_{ij}|x_{ij})$ is illustrated in the figure \ref{f_z}.
    
\begin{figure}[htbp]\centering
	\includegraphics[trim=0cm 1cm 12cm 5cm, clip, width=6cm]{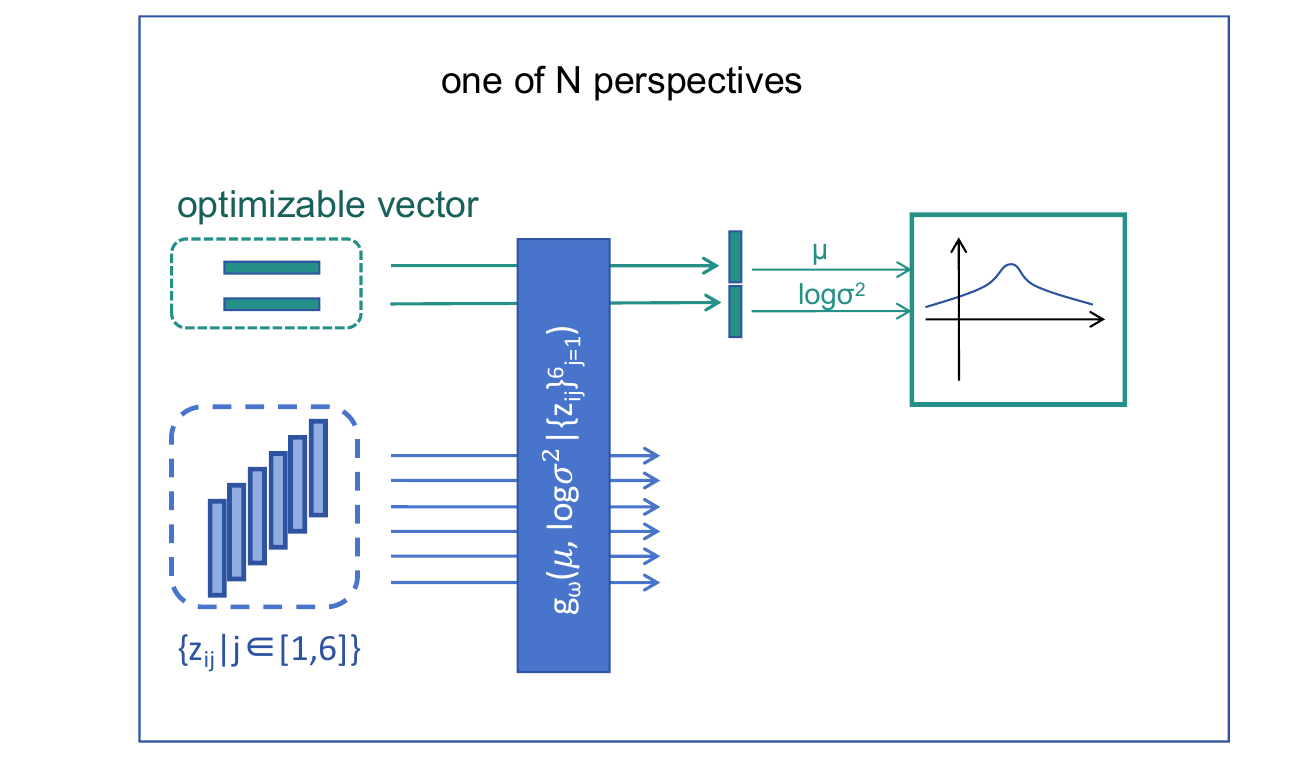}
	\caption{Parameterizes process of $p_i'(z|y)$}
\label{g_omega}
\end{figure}

\begin{figure}[htbp]\centering
	\includegraphics[trim=0cm 0cm 2cm 0cm, clip, width=6cm]{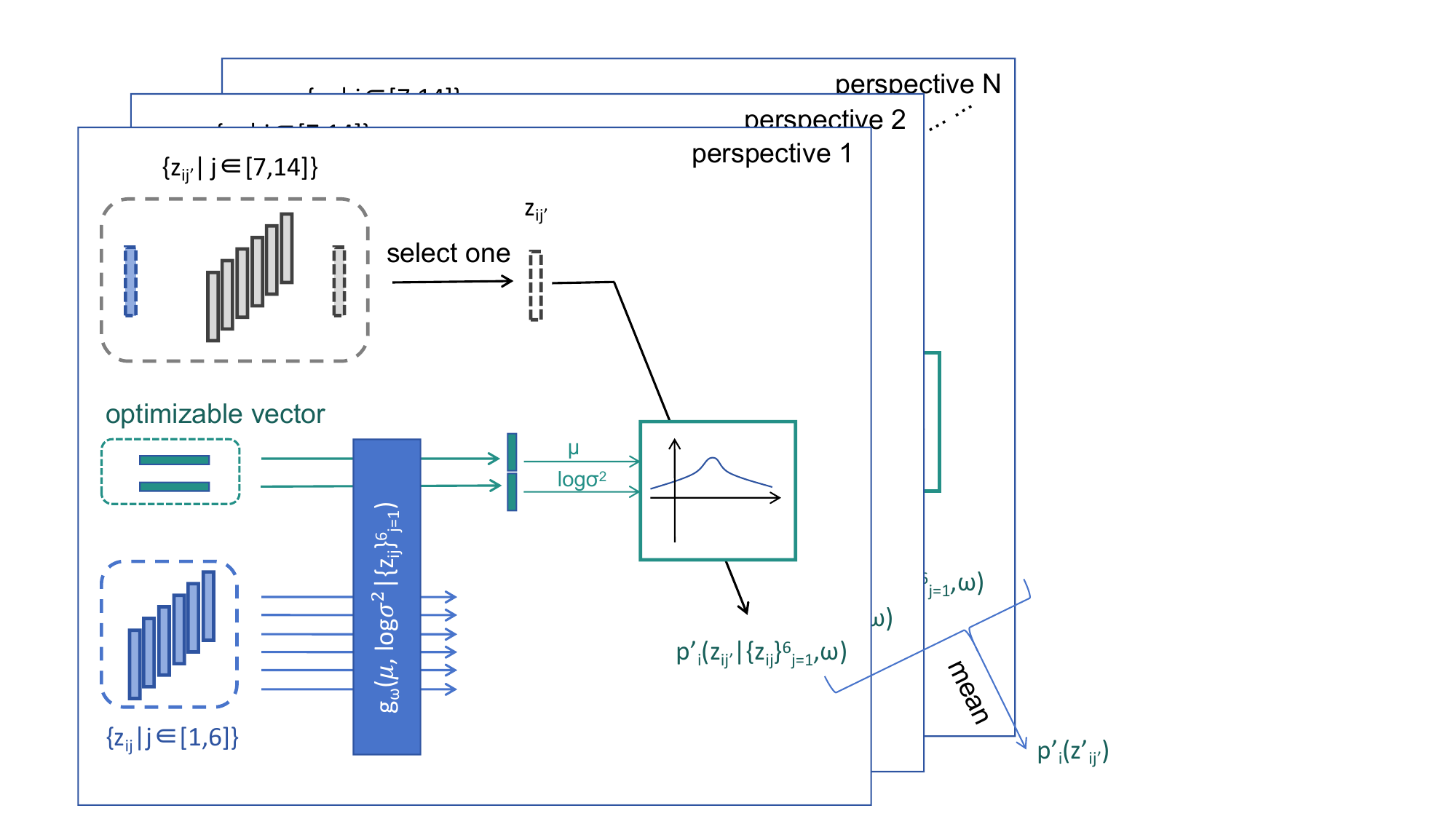}
	\caption{The calculation process of the final probability}
\label{final_score}
\end{figure}

\begin{figure}[htp]\centering
	\includegraphics[width=8.5cm]{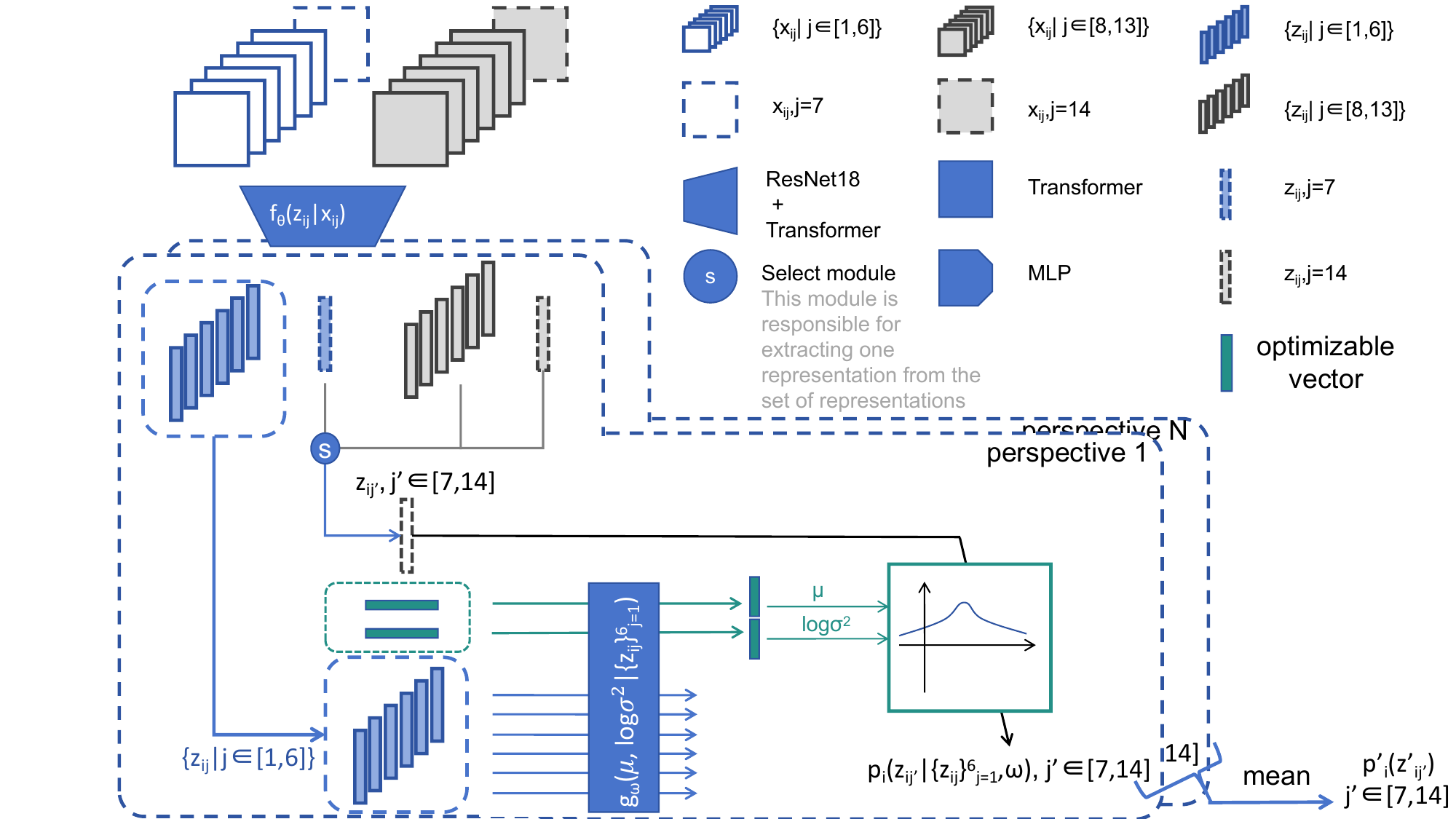}
	\caption{Feedforward process of PMoC.}
\label{g_mu_logvar}
\end{figure}

The network $g_\omega (\mu_i , \sigma_i |\{z_{ij}\}^6_{j=1})$ is much more of its name. It encodes the mean and variance of the $i$-th primary group distribution based on the latent representations of the $i$-th primary group samples, and $i$ still represents the problem index in Bongard-Logo. Specifically, we uniformly process each perspective $z_{ij}$ within $z'_{ij}$. Under a single perspective, we treat $\{z_{ij}\}^6_{j=1}$ as six tokens and, after appending two optimizable vectors along with positional embedding, input them into $g_\omega (\mu_i , \sigma_i |\{z_{ij}\}^6_{j=1})$ with a Transformer-Encoder backbone for computing attention results. We extract two optimizable vectors from the attention results, serving as the mean $\mu_i \in R^d$ and logarithm of variance $\log(\sigma_i ^2) \in R^d$ for a multivariate Gaussian distribution. The encoding distribution pattern of module $g_\omega (\mu_i , \sigma_i |\{z_{ij}\}^6_{j=1})$ follows a similar approach to that of VAE\cite{VAE}; however, it does not incorporate variational inference in its design. The aforementioned process parameterizes $p_i'(z|y)$ as a multivariate Gaussian distribution and outlines the computation of $g_\omega (\mu_i , \sigma_i |\{z_{ij}\}^6_{j=1})$ and this process can be expressed as Figure \ref{g_omega}.
We feed 8 vectors $\{z_{ij'}\}^{14}_{j'=7}$ into this distribution for calculating 8 logical probabilities $\{{p}_{i}'(z_{ij'}|\{z_{ij}\}^6_{j=1}, \omega)\}_{j'=7}^{14}$ in one perspective. In this paper, we calculate the final probability ${p}_{i}'(z'_{ij'})$ by averaging the logical probabilities ${p}_{i}'(z_{ij'}|\{z_{ij}\}^6_{j=1}, \omega)$ obtained from all perspectives. The calculation process of the final probability can be expressed as Figure \ref{final_score}.
By utilizing the cross entropy loss function, we constrain the final probability ${p}_{i}'(z'_{i7})$ to approach 1 while ensuring other probabilities $\{{p}_{i}'(z'_{ij})\}_{j=8}^{14}$ are close to 0. Figure \ref{g_mu_logvar} illustrates in detail the forward process of the PMoC. However, this approach did not yield satisfactory results.

We speculate that the reason for the ineffectiveness of the aforementioned method lies in the unknown true form of the distribution $p_{i}'(z|y)$, which may be a complex mixed Gaussian distribution. Therefore, it may not be reasonable to parameterize $p_{i}'(z|y)$ as a multivariate Gaussian distribution and compute the probability $p_{i}'(z_{ij'}|y)$. Additionally, the formula for calculating the logical probability of a vector $z$ belonging to the parameterized distribution is provided as follows:
\begin{align}
    \log(p(z ,\mu , \sigma )) = \frac{{(z_{m} - \mu _{m})^{2}}}{{2\sigma^{2}_{m}}} - \log(|\sigma_{m}|) - \frac{1}{2}\log(2\pi)\label{normal_distribution}
\end{align}
Where $m$ represents the dimension index of the vector. $\mu$ represents the mean of the multivariate Gaussian distribution, $\sigma$ represents the standard deviation of the multivariate Gaussian distribution, and $z$ is the representation of some unknown data. It is evident that optimizing the network using the given formula (\ref{normal_distribution}) as the loss function presents challenges, as optimizing the above equation can be indirectly viewed as optimizing the Mean Squared Error (MSE) loss between vectors. Moreover, these vectors have not undergone normalization or squashing, making it difficult to optimize the network from the perspective of implementation.

To counter such ineffectiveness, we decide to train a deep network to directly fit probability $p_{i}'(z_{ij'}|y)$. We argue that, rather than laboriously analyzing and fitting a distribution, it is more efficient to directly compute the probability of a sample belonging to that distribution. Inevitably the network's formulation has been transformed from $g_\omega (\mu_i , \sigma_i |\{z_{ij}\}^6_{j=1})$ to $g_\omega (p_{i}'(z_{ij'}|y) |\{z_{ij}\}^6_{j=1}, z_{ij'})$, thereby shifting the problem.
Different from previous approaches $g_\omega (\mu_i , \sigma_i |\{z_{ij}\}^6_{j=1})$ that fit the distribution  $p_{i}'(z|y)$, this paper employs a Transformer-Encoder to fit the probability  $p_{i}'(z_{ij'}|y)$ directly, which enables the direct computation of the probability that $z_{ij'}$ belongs to the distribution $p_{i}'(z|y)$.

\begin{figure}[htp]\centering
	\includegraphics[trim=0cm 3cm 5cm 3cm, clip, width=6cm]{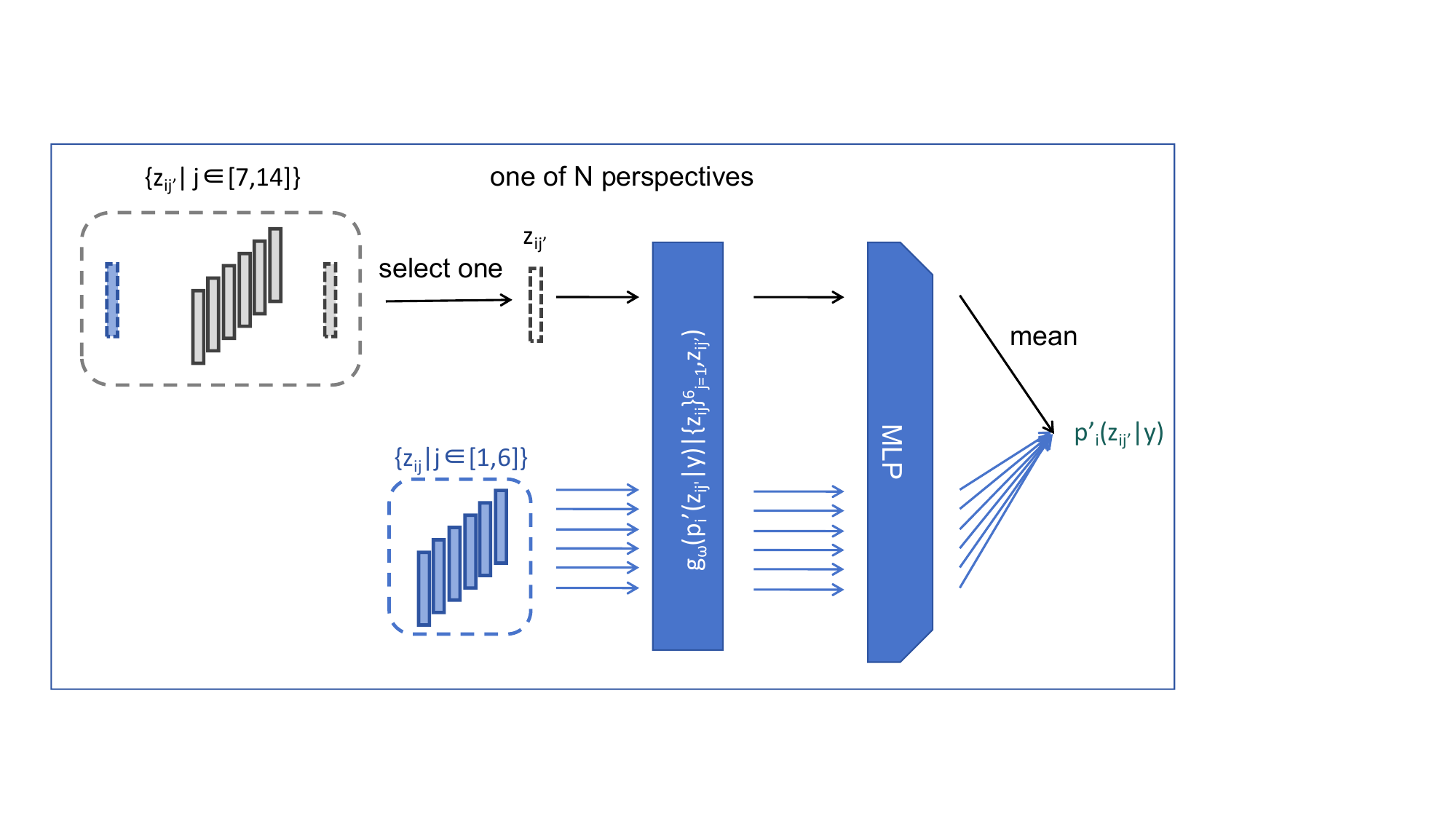}
	\caption{Feedforward process of $g_\omega (p_{i}'(z_{ij'}|y) |\{z_{ij}\}^6_{j=1}, z_{ij'})$.}
\label{new_g_omega}
\end{figure}
\begin{figure}[htbp]\centering
	\includegraphics[trim=0cm 0.2cm 0cm 2cm, clip, width=6cm]{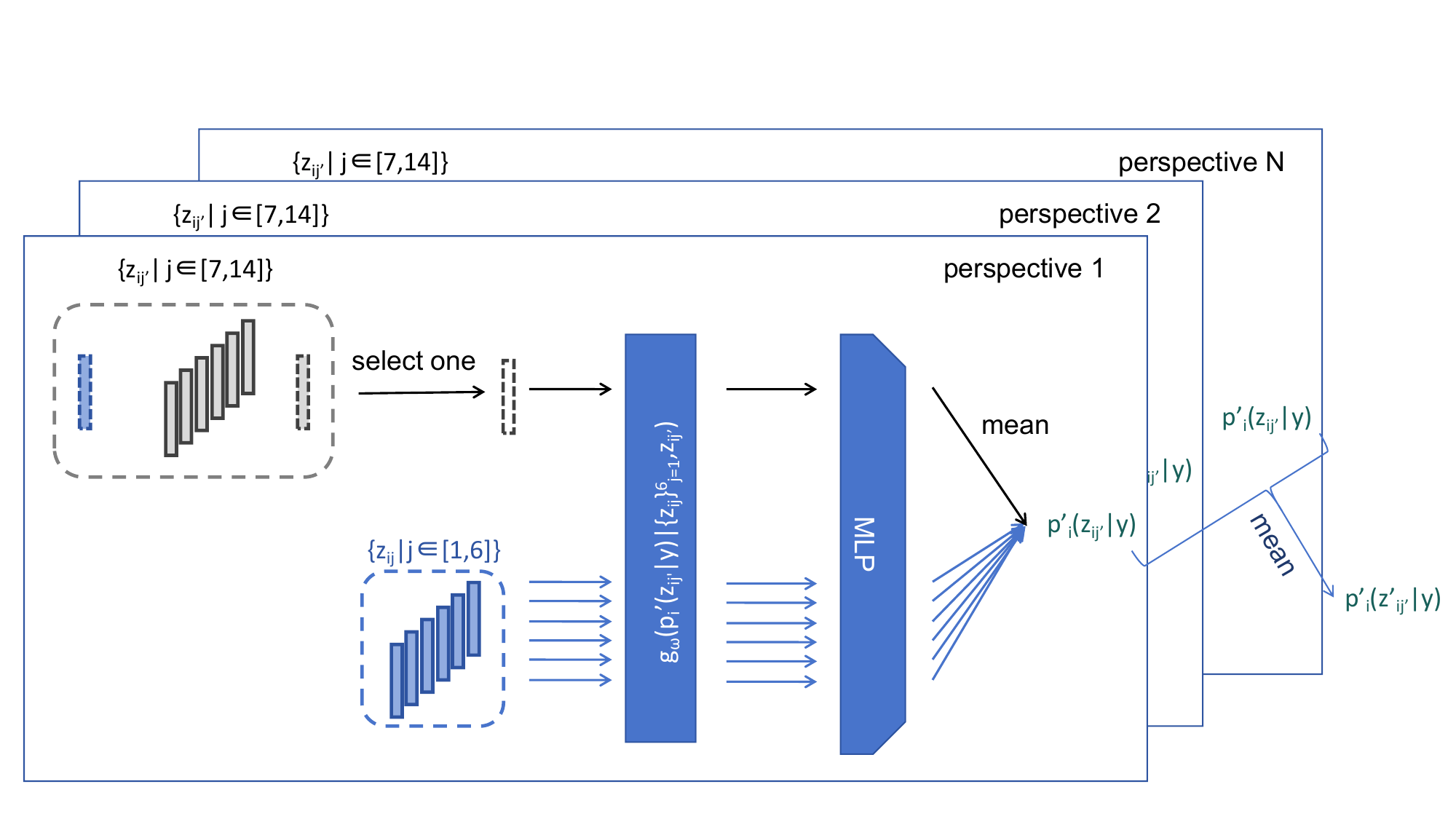}
	\caption{The new calculation process of the final probability}
\label{new final_score}
\end{figure}
\begin{figure}[htp]\centering
	\includegraphics[width=8.5cm]{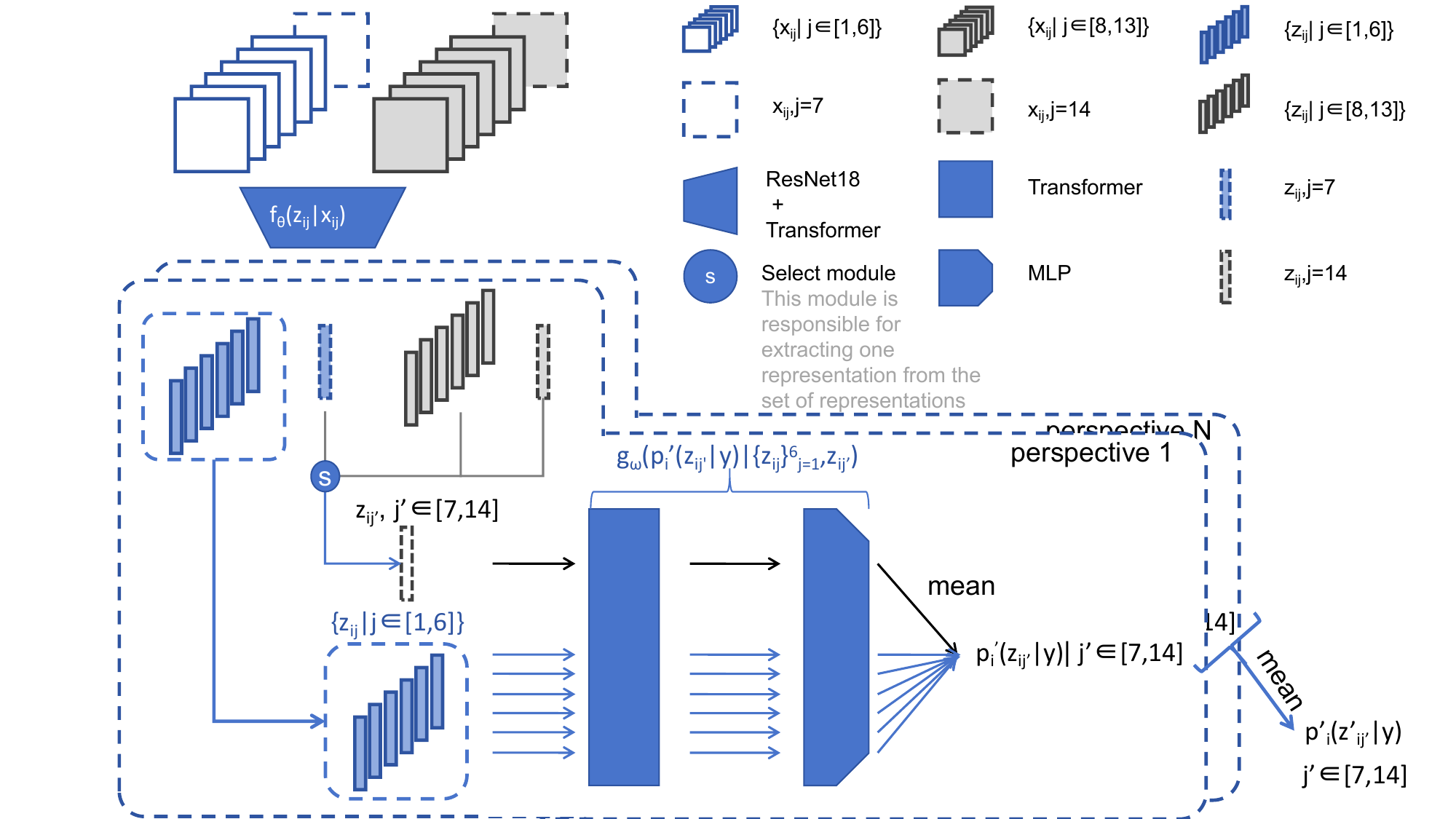}
	\caption{Feedforward process of PMoC 2.0 ver.}
\label{g_z}
\end{figure}

For the implementation on the new form of $g_\omega(\cdot|\cdot)$, the function $g_\omega(\cdot|\cdot)$ maintains the egalitarian approach in processing each perspective $z_{ij}\in R^{d}$ within the outcomes $z'_{ij}\in R^{n\times d}$ encoded by $f_\theta(z_{ij}|x_{ij})$. We amalgamate $\{z_{ij}\}^6_{j=1}$ with one of the pending representations $\{z_{ij'}\}^{14}_{j'=7}$, resulting in a set of seven tokens. Following the embedding of positional encoding, these tokens are then systematically fed into a Transformer-Encoder, aimed at yielding 7 attention results.
All 7 attention results are mapped to 7 probabilities by a MLP. The mean of the mapped outcomes is considered as the logical probability $ {p}_{i}'(z_{ij'}|y)$ corresponding to the specified perspective. This constitutes the feedforward process of the $g_\omega (p_{i}'(z_{ij'}|y) |\{z_{ij}\}^6_{j=1}, z_{ij'})$. This feedforward process can be expressed as Figure \ref{new_g_omega}.
Each pending representation $z_{ij'}$ within $\{z_{ij'}\}^{14}_{j'=7}$ receives a corresponding probability output $p_{i}'(z_{ij'}|y)$. 
Each logical probability stored in $\{{p}_{i}'(z_{ij'}|y)\}_{j'=7}^{14}$ is obtained based on a single perspective. By averaging the logical probabilities across all perspectives, we derive the final probability values $\{ p_{i}'(z'_{ij'}|y)\}_{j'=7}^{14}$. The new calculation process of the final probability can be expressed as Figure \ref{new final_score}.
In this paper, we still optimize the model using the cross-entropy loss function. It is worth mentioning that the essence of the final version of PMoC lies in its function as a distribution distance metric, thereby necessitating the satisfaction of Lipschitz continuity. To ensure this, we have imposed spectral normalization constraints on the parameters of PMoC. Figure \ref{g_z} visually depicts the process of PMoC.

In summary, this paper makes two compromises for PMoC. Firstly, we acknowledge that compared to $p_i'(z)$, $q_i'(z)$ is more likely to be a complex distribution. For instance, $p_i'(z)$ may follow a Gaussian distribution, while $q_i'(z)$ may adhere to a mixed Gaussian distribution. Therefore, estimating these two distributions and utilizing them to solve the Bongard-Logo problem is challenging to achieve excellent performance. Accordingly, we shift our focus to estimating the distribution $p_i'(z)$ and utilize the probability of a representation belonging to this distribution to determine whether it is a primary or auxiliary group representation. Secondly, due to the challenges associated with optimizing the model from the perspective of estimating $p_i'(z)$, this paper opts to directly compute the probability of a representation belonging to the distribution $p_i'(z)$. 


Despite the apparent disadvantage of probabilistic clustering reasoning solvers, we persist in designing and developing a Bongard-Logo solver in the form of a probabilistic model. Our aim is not to rival the most advanced solutions available, but rather to contribute to the diversity of deep learning research outcomes.
We are motivated by the goal of fostering innovation and exploration beyond the mainstream, encouraging a wider range of approaches and perspectives in the field. We are confident that a probabilistic clustering approach can offer more intuitive probability assessments for clustering decisions, which could be critically important in certain analogous fields.

\section{Experiment}

In this section, we conduct experiments on the model proposed in this paper.

\subsection{Experiment on Bongard-Logo}
All our experiments on Bongard-Logo are implemented in Python using the PyTorch\cite{
Pytorch} framework. In the experiments conducted for PMoC and SBSD, the Adam\cite{ADAM} optimizer was selected with a learning rate set to 0.001, a learning rate decay of 0.995, a weight decay of 0.0001, and a batch size of 40 for PMoC and SBSD. 

We conducted experiments on the Bongard-logo dataset, utilizing SBSD, PMoC. The experimental results are documented in Table \ref{Bongard_Results}. Due to the propensity of the serial convolutional scanning mechanism and attention mechanism to experience attention collapse when combined, we adopted the parameters of the SBDC model as pre-trained parameters for the PMoC convolutional layer. 

\begin{table}[h]
\caption{Reasoning Accuracies of PMoC on Bongard-logo.}
\label{Bongard_Results}
\centering
\resizebox{\linewidth}{!}{
\begin{tabular}{ccccccc}
\toprule
&\multicolumn{5}{c}{ Accuracy(\%)}& \\
\cmidrule{2-6}
Model&Train& FF&BA&CM&NV \\
\midrule
SNAIL&59.2&56.3&60.2&60.1&61.3\\
\midrule
ProtoNet&73.3&64.6&72.4&62.4&65.4\\
\midrule
MetaOptNet &75.9&	60.3&	71.6&	65.9&	67.5\\
\midrule
ANIL & 69.7 & 56.6 & 59.0 & 59.6 & 61.0\\
\midrule
Meta-Baseline-SC & 75.4 & 66.3 & 73.3 & 63.5 & 63.9 \\
\midrule
Meta-Baseline-MoCo & 81.2 & 65.9 & 72.2 & 63.9& 64.7 \\
\midrule
WReN-Bongard & 78.7 & 50.1 & 50.9 & 53.8 & 54.3 \\
\midrule
\midrule
SBSD&83.7&75.2&91.5&71.0&74.1\\
\midrule
PMoC&92.0&90.6&97.7&77.3&76.0\\
\bottomrule
\end{tabular}
}
\end{table}

In Table \ref{Bongard_Results}, it is evident that addressing the Bongard-Logo problem using a probabilistic model approach, namely PMoC, is both straightforward and effective. Furthermore, SBSD exhibits stronger outcomes when compared to previous models. 

Finally, we postulate that the significance of PMoC, a network primarily structured around the Transformer-Encoder, may be further emphasized when dealing with larger datasets. ViT \cite{ViT} explicitly explores the importance of data volume for Transformer-based visual models, whereas RS-Tran \cite{RS} focuses on the significance of data quantity in addressing visual abstract reasoning challenges. Therefore, this paper attempts to conduct experiments on an expanded Bongard-Logo dataset to explore the further capabilities of PMoC. However, due to the limited availability of additional Bongard-Logo data for expanding the dataset, this paper uses the data augmentation techniques to achieve the objective. To expand the Bongard-Logo problems, we employ data augmentation techniques such as random rotating by integer angles ($90^\circ$, $180^\circ$, $270^\circ$) and random horizontal/vertical flipping. These augmentations are applied to implicitly increase the data volume without altering the essence of Bongard-Logo as a few-shot learning task or explicitly changing the level of reasoning difficulty. We anticipate that this augmentation approach will enable PMoC to unleash its greater potential. The results is illustrated in Table \ref{augmented_Bongard_Results}.

\begin{table}[h]
\caption{Reasoning Accuracies of PMoC on Augmented Bongard-logo.}
\label{augmented_Bongard_Results}
\centering
\resizebox{\linewidth}{!}{
\begin{tabular}{ccccccc}
\toprule
&\multicolumn{5}{c}{ Accuracy(\%)}& \\
\cmidrule{2-6}
Model&Train& FF&BA&CM&NV \\
\midrule
PMoC&94.5&92.6&98.0&78.3&76.5\\
\bottomrule
\end{tabular}
}
\end{table}

This paper argues that the adopted data augmentation methods do not explicitly alter the core reasoning difficulty or logical complexity of the Bongard-Logo problem. These augmentation techniques, from an intuitive perspective, neither add nor remove any preexisting concepts. Even if these data augmentation methods implicitly introduce additional and unknown concepts to the Bongard-Logo dataset, it is important to note that such concepts are not present in the test set, as no data augmentation has been applied to it. Furthermore, they do not provide guidance for generalization issues such as NV and CM, as the solutions to these problems are not directly reflected in the HD problem after applying such augmentations. Additionally, the data augmentation approach employed in this study is uniformly applied to all 14 images within each Bongard-Logo instance, maintaining the nature of few-shot learning (i.e., without increasing the number of images within an instance) and without introducing additional priors (such as rotational and flipping invariance of individual images within Bongard-Logo instances). These augmentation methods indeed enrich the expressions of concepts at the pixel level, which is precisely the challenge we anticipate for PMoC through data augmentation and expanded data volume.
Enabling multiple expressions for concepts is precisely the original intention behind the establishment of the Bongard-Logo dataset\cite{Bongard2}.

\section{Conclusion}

The characteristics of deep networks as black-box intelligent systems, coupled with their optimization methods based on steepest gradient descent, create a gap between the theory and implementation of deep networks. In this paper, we report the design process of PMoC, which represents a ``compromise" and ``indirect" approach to neural network architecture design philosophy. By employing a unique probabilistic modeling technique, PMoC cleverly circumvents many of the theoretical and practical discrepancies associated with deep learning. This allows PMoC to offer a probabilistic-model-based solution for visual clustering problems. Its effectiveness has been validated on the renowned clustering reasoning problem: Bongard-Logo. This paper advocates for the adoption of similar indirect strategies and compromise approaches when addressing similar visual clustering problems or tasks requiring the estimation of unknown distributions. 


\end{document}